# UT-Net: Combining U-Net and Transformer for Joint Optic Disc and Cup Segmentation and Glaucoma Detection

Rukhshanda Hussain*, Hritam Basak*

*Abstract*—Glaucoma is a chronic visual disease that may cause permanent irreversible blindness. Measurement of the cup-to-disc ratio (CDR) plays a pivotal role in the detection of glaucoma in its early stage, preventing visual disparities. Therefore, accurate and automatic segmentation of optic disc (OD) and optic cup (OC) from retinal fundus images is a fundamental requirement. Existing CNN-based segmentation frameworks resort to building deep encoders with aggressive downsampling layers, which suffer from a general limitation on modeling explicit long-range dependency. To this end, in this paper, we propose a new segmentation pipeline, called UT-Net, availing the advantages of UNet and transformer both in its encoding layer, followed by an attention-gated bilinear fusion scheme. In addition to this, we incorporate Multi-Head Contextual attention to enhance the regular self-attention used in traditional vision transformers. Thus low-level features along with global dependencies are captured in a shallow manner. Besides, we extract context information at multiple encoding layers for better exploration of receptive fields, and to aid the model to learn deep hierarchical representations. Finally, an enhanced mixing loss is proposed to tightly supervise the overall learning process. The proposed model has been implemented for joint OD and OC segmentation on three publicly available datasets: DRISHTI-GS, RIM-ONE R3, and REFUGE. Additionally, to validate our proposal, we have performed exhaustive experimentation on Glaucoma detection from all three datasets by measuring the Cup to Disc Ratio (CDR) value. Experimental results demonstrate the superiority of UT-Net as compared to the state-of-the-art methods.

*Index Terms*—Glaucoma, Medical Image Analysis, Optic Cup Segmentation, Optic Disc Segmentation, Transformer Network

## I. INTRODUCTION

GLAUCOMA is the second largest cause of blindness worldwide and the leading cause of irreversible blindness [1]. It is caused by excessive fluid pressure inside the inner portion of the eye, also known as intraocular pressure (IOP). This causes the optic cup (OC) to grow in shape than the optic disc (OD), known as *cupping*, and may also thicken the wall of retinal nerve fibers (RNFL). Since the visual loss due to glaucoma is irreversible, proper screening, as well as

R. Hussain is with Dept. of Electrical Engineering, Jadavpur University, Kolkata, India
H. Basak is with Dept. of Computer Science, Stony Brook University, New York, USA
Corresponding author; hritambasak48@gmail.com *
The authors contributed equally.

diagnosis is essential for preserving vision. One of the widely used screening methods is the optic nerve head (ONH) assessment, which is the binary classification between healthy and diseased subjects. However, the large-scale assessment of glaucoma using ONH by expert clinicians is time-consuming, laborious, and practically impossible for population-wise screening.

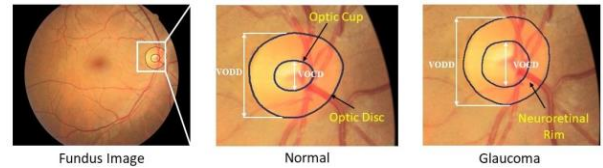

Fig. 1: Structure of retinal fundus: the region enclosed by the outer ring is the optic disc, whereas the region enclosed inside the smaller ring is the optic cup, and the region between them is the neuroretinal rim. The optic cup gets larger in case of glaucoma. The figure is taken from the open-source DRISHTI-GS dataset [2].

Some clinical methods such as rim-to-disc area ratio (RDAR), cup-to-disc ratio (CDR), and disc diameter are proposed for population-wise screening. Among these CDR is quite popular and widely used by physicians. CDR is equivalent to the ratio between vertical optic cup diameter (VOCD) and vertical optic disc diameter (VODD), as shown in Figure 1. In glaucoma, the OC region grows in size and a higher CDR indicates a higher risk of having glaucoma and vice versa. Accurate segmentation of OD and OC is therefore of utmost importance to measure the CDR and assess the condition of a patient. Measuring glaucoma manually, although considered the gold standard in clinical practice, is prone to inter-observer variability and is not always feasible due to a lack of expert clinicians. On the other hand, computer-aided methods are robust, cost-effective, and require the least human intervention, thereby making them free from human error [3]. Some of the methods utilize optical coherence tomography (OCT) images for automatic segmentation of OD and OC and retinal layer boundaries [4], however availing of OCT is costly and difficult. Therefore, CDR is considered the most reliable and widely used method, for which segmentation of OC and OD is extremely important.

Deep learning methods, specifically Convolutional Neural Networks (CNNs) have been widely used recently in several image segmentation tasks. [5], [6]. Despite their unparalleled performance in medical image segmentation, CNNs fail to model long-range relations due to the intrinsic localization of convolution operations [7]. Therefore, they yield poor performance in cases having large inter-case variability. To address this limitation, attention mechanisms have been proposed, without making significant changes in the CNN backbones, that produced superior performance than the former ones. The convolutional operation employs a particular kernel size which decides the field of reception. Nevertheless, the nature of such operations remains localized, i.e., remain restricted to the area of intersection between the image and the kernel of a particular size. Transformers, on the other hand, utilize dispense convolution and attention mechanisms and unlike CNN, they can model global contexts alongside their superior transferability power for downstream tasks under huge-scale training. They have been widely adopted for multiple aspects including medical image segmentation [8]. A transformer considers an image to be equivalent to $N \times N$ words in Natural Language Processing (NLP). Therefore, the input image is split into patches that are linearly embedded with positional encodings. The obtained sequence vector is then fed to the transformer layers. Owing to the long-term memory of the transformers because of the attention mechanism, such models can 'attend' to all previously generated tokens. Hence it enables the encoder to map sequences into an abstract continuous representation that holds the learned information for that entire sequence, thereby establishing the global dependency of acquired features. However, segmentation frameworks solely based on transformers like SETR [9] fail to produce satisfactory results because they lack spatial inductive bias in modeling local information, and hence, may not be useful for medical image analysis.

Several deep learning-based segmentation pipelines have been proposed in the literature that is capable of learning complex features. For instance, Sevatopolsky et al. [10] developed a modified U-Net-based light framework having a lesser number of filters in convolutional layers. Tabassum et al. [11] proposed an encoder-decoder-based architecture on the SegNet backbone with dense connections introduced in the encoder for OD and OC segmentation. However, a common problem with U-Net alike architectures is that consecutive convolution operations followed by pooling reduce the feature resolution that affects the learning of abstract feature representations. Authors of the work [12] introduced deconvolutions in the feature decoder to overcome the stated problem. Additionally, dense atrous convolutions were incorporated to encode high-level semantic maps in the downsampling half. In [13] the authors proposed an automated CNN-based U-shaped encoder-decoder architecture consisting of residual skip connections consisting of two shortcut paths compensating for the spatial information loss due to the pooling operation. However, despite leveraging the skip connections some data is still being lost in the pooling which is further transferred to the decoding half affecting the image reconstruction to its original resolution. On the other hand, maintaining the same resolution throughout the convolutions can be computationally expensive.

Transformers, on the other hand, try to extract the features from smaller patches of original resolution thereby compensating to some extent for the features lost. He et al. [14] proposed adding an extra branch to Residual CNNs (RCNN) to predict the object mask in parallel to the existing RCNN branch, facilitating the bounding box prediction to obtain a dice score of $90.45$ on the REFUGE dataset. The boundaries of the OC are often spatially sparse, which makes the segmentation of the OC to be relatively more difficult than that of OD. To address this difficulty Liu et al. [15] proposed an atrous CNN framework consisting of a pyramid filtering module to obtain multiscale spatial aware features and the model was tested on the DRISHTI-GS dataset. Ngo et al. [16] considered the retinal layer segmentation task as a regression problem, considering the intensity, gradient, and adaptive normalized intensity score (ANIS) of OCT images as features. Shankaranarayana et al. [17] proposed a novel dilated residual inception block for multiscale feature extraction simultaneously integrated with a fully convolutional guided network. An improved model, called U-Net++, was proposed by Tulsani et al. [18] with a custom loss function to mitigate the class imbalance problem which occurs due to the small size of the ONH. A common line of thought in the above works would be enhancing the CNN framework to extract finer localized features and eventually improving the segmentation performance. Additionally, [17] solidifies the effectiveness of multi-scale feature extraction with CNNS. based on this we incorporate a modified receptive field block with dilated convolutions in each of the skip connections. However, in order to mitigate the issue of finding the optic cup region within the disc we rely on using global contextual information extracted from image patches via a transformer encoder.

Recently, self-attention-based models, like transformers [19] have emerged to be extremely efficient in capturing global contextual information. Chen et al. [7] proposed a hybrid model incorporating a transformer encoder in the bottleneck of the U-Net framework. Another approach for integrating transformers with a CNN-based network was proposed in [20], replacing the downsampling path of U-Net completely with a pure transformer encoder to learn sequential representations in the input image. While this scheme works well, low-level spatial information detrimental to generating accurate segmentation maps is not efficiently captured with a stand-alone transformer encoder. A symmetric U-shaped pure transformer network with skip connections was proposed by Cao et al. in [21]. In addition, the conventional transformer blocks consisted of two successive window-based Multi-Head Attention modules. Nonetheless, a pure transformer network being too deep is difficult to converge. A possible solution to it could be the application of a bottleneck [21] that is capable of learning deep features. Inspired by this, in our work, we retain the long-skip connections of existing U-Net itself in order to retain and propagate features that may be lost due to bottleneck layers between the encoder and the decoder sections, and can be used for better reconstruction of the image by the decoder [22].

Most of the existing methods in the literature fail to generate accurate segmentation maps of OC regions from fundus images, resulting in misleading CDR measurement. The reason is the small dimension of OC and minor gradient change near the OC boundary region. Convolutional networks like U-Net, although the focus on the global context, cannot identify these local changes, hence failing to localize OC regions accurately. Transformers, on the other hand, consider the sequence of pixels in the form of patches, and hence it is easier for transformers to identify pixel-wise gradient information in the small patches. Besides, the proposed multi-head contextual attention in our transformer produces features containing contextually rich key representations. The self-attention also holds some memory of the learned spatial gradient information and utilizes the knowledge in later stages.

Driven by the aforementioned speculations, we propose the UT-Net, in this paper, utilizing the advantages of transformer and CNN both, for joint OD and OC segmentation and glaucoma detection. We also measure the CDR using segmentation maps generated from the network. The proposed model can essentially be leveraged in CAD systems to assist medical professionals in their diagnosis and eventual treatment. Furthermore, robust segmentation outputs might help clinicians in measuring the severity of glaucoma and monitoring the patient's condition. The contributions of this work can be listed as follows:

1) We propose a novel encoder-decoder architecture that contains a transformer and U-Net as two of its encoders in two parallel branches. This multi-encoder pipeline helps the network compensate for the intrinsic localization problem of convolution operations using the self-attention mechanism of the transformer and also helps the network to extract more discriminative features.
2) A self-attention module in vision transformers generates attention matrix $A$ from isolated query-key pairs at a particular spatial location and hence neglects the global context available in the surrounding. To overcome this problem, we introduce Multi-Head Contextual Self attention to enhance the preliminary attention matrix.
3) The extracted features from U-Net and transformer encoders are merged through a bilinear fusion module that utilizes spatial and channels attention operations on the features in two parallel branches, followed by their intelligent fusion using the Hadamard product.
4) The skip connections of U-Net are replaced with multiscale context extractors for the synthesis of receptive field information at multiple levels from the encoders.
5) The proposed UT-Net model has been evaluated on three publicly available retinal fundus image datasets: DRISHTI-GS, RIM-ONE R3, and REFUGE for joint OD and OC segmentation, and Glaucoma detection. When compared to the state-of-the-art methods, UT-Net outperforms them by a significant margin.

## II. MATERIALS AND METHODS

In this section, we elucidate the overall workflow of the proposed UT-Net model and describe its components. The architecture of the proposed UT-Net is shown in Figure 2.

### A. U-Net-based Encoder

The first branch of the downsampling path in the proposed architecture consists of a U-Net [22] encoder as shown in Figure 2. This branch contains 5 steps, each of them consisting of a 3 × 3 convolution operation, followed by batch normalization. For incorporating non-linearity into the model, each output of the batch normalization is passed through a Rectified Linear Unit (ReLU) activation followed by a 2 × 2 MaxPooling operation. This downsampling path progressively extracts image features and increases these feature dimensions in every layer. Thus, ultimately a high-dimensional feature representation with high semantic information is obtained at the final layer. Skip connections are introduced, enabling the fine-grained information from the encoder to be recovered in the upsampling half of the network.

### B. Transformer Encoder

A transformer encoder follows stacks of self-attention, which is as a query and key-value pair with point-wise fully connected layers. The output is the weighted sum of the values and this weight is computed by a compatibility function of the corresponding query. Sequential operations of CNNs often fail to capture the long-range dependency information which is essential in the case of semantic segmentation [19]. To alleviate this problem, a transformer encoder is introduced that effectively captures the global contextual information. To make the paper self-contained, we briefly describe the transformer encoder network here.

Firstly, an input image, $Z \in \{H \times W \times C\}$ is divided evenly into $N = H/P \times W/P$ patches The grid of patches is flattened into a sequence that is passed through the linear embedding layer having an output dimension of $D_l$. Learnable positional embeddings of matching dimensions are added to the raw embedding sequence, $e_b \in \mathbf{F}^{N \times D_l}$ for utilization of the spatial prior. The embedding sequence, $Z_l$ thus obtained acts as the input to the transformer encoder consisting of 12 layered Multi-Head Self-attention (MSA) [23] followed by multi-layer perceptron (MLP) [24]. Unlike the regular transformers, we utilize Multi-Head Contextual Self attention (MCSA) instead of MSA.

Multi-Head Contextual Self-attention: A self-attention module in vision transformers is generally applied on 2-D feature maps which obtains the attention matrix, $A$ from isolated query-key pairs at a particular spatial location. However, the global context available in the neighboring keys is often overlooked, which is detrimental to medical datasets owing to its high cross-image similarity. To alleviate this, we incorporate the contextual self-attention block as shown in Figure 2.

The self-attention (SA) mechanism of a transformer updates the state of each patch by global aggregation of features and can be formulated as in Equation 1:

$$SA(Z) = Z + softmax\left(\frac{ZW_q(ZW_k)^T}{\sqrt{d}}\right)(ZW_v) \quad (1)$$

where $Z$ is the input given to the self-attention is a triplet of a query, a key, and a value such that query = $ZW_q$, key = $ZW_k$, and value = $Z$, where, $W_v$, $W_q$, and $W_k$ are learnable parameters. Unlike the 1×1 convolutions adopted in vision transformers for encoding every individual key, MCSA employs a group $k \times k$ convolution on the neighboring keys to construct a spatial $k \times k$ grid containing contextually rich key representations. Thus, the newly learned keys, $K' \in R_{H \times W \times C}$ now contain the contextual information from the local adjacent keys. Let $K'$ be a consistent representation of $X$ in space. The key, $K'$ and query, $Q$ are concatenated, and the attention matrix is obtained after two sequential convolutional operations where learnable weights $W_\theta$ is with ReLU activation and $W_\phi$ is without activation. The attention matrix can be formulated as in Equation 2

$$A(Z) = [QK'^T]W_\theta W_\phi \quad (2)$$

For a particular head, the attention matrix at every spatial location of $A$ is learned by leveraging the contextual query-key pair instead of the localized query-key pair. The self-attention output is thereby enhanced using statically mined $K'$. The attentive feature map $K''$ is obtained by aggregating the values as in regular SA expressed as in Equation 3:

$$K'' = ZW_v \odot A(Z) \quad (3)$$

The MCSA consists of $m$ such SA operations to project a concatenated output which is then transformed by a residual skip operation. The output can be expressed as in Equation 4:

$$output = MCSA(Z) + MLP(MCSA(Z)) \quad (4)$$

Finally, the output undergoes layer normalization and is passed on to the bilinear fusion block. The transformer encoder architecture is shown in Figure 2.

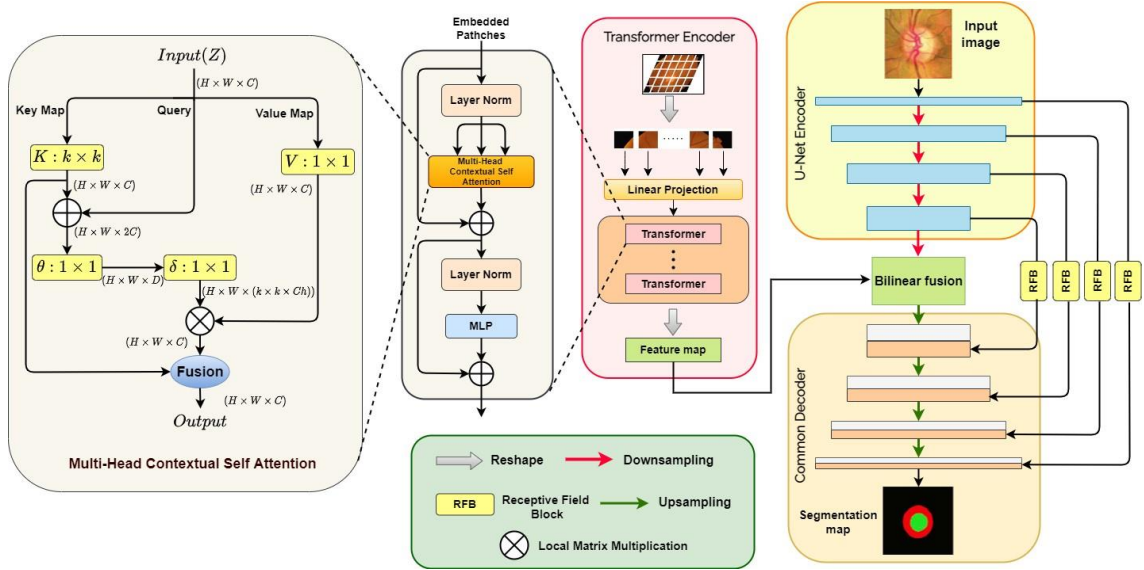

Fig. 2: Overall architecture of the proposed UT-Net, consisting of two independent encoder branches: U-Net and Transformer, where the transformer receives patch-based images whereas the U-Net receives the whole image itself. Features extracted from both these encoders are fused through a bilinear fusion block, followed by a common decoder path that produces the final segmentation map. Modified RFB blocks are used for the effective utilization of context information in the encoded features

### C. Bilinear Fusion Block

For an effective combination of the features obtained from the two encoder branches, a bilinear fusion block is incorporated which consists of a self-attentive multi-modal fusion mechanism as shown in Figure 3. The fusion is performed through a series of linear operations on features obtained from two encoders in two parallel branches and finally merging them. Channel attention is used for promoting global information from the transformer branch. To filter out any noises that might be present in the low-level features extracted from the U-Net branch, a spatial attention operation is used. Finally, a Hadamard product is taken between the weights obtained from the two respective branches thus modeling the fine-grained interaction between these two sets of features.

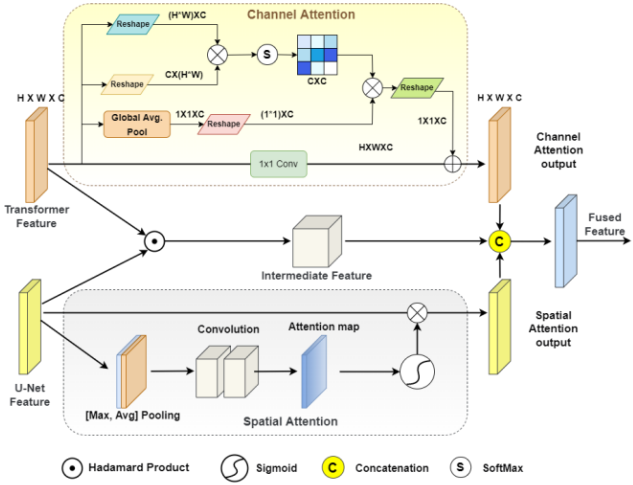

Fig. 3: Bilinear fusion scheme of features extracted from two different encoder branches.

Channel Attention: In each of the channels, the features learned by the filters operate on a local receptive field resulting in inefficient exploration of features outside the local region. To mitigate the problem, the global spatial information from the transformer branch is squeezed into a channel descriptor by average pooling, generating channel-oriented statistics. First, the input feature $F \in R^{H \times W \times C}$ is reshaped into intermediate features $F_1 \in R^{(H*W) \times C}$ and $F_2 \in R^{C \times (H*W)}$, followed by √ their multiplication, and division by a factor of $C$. Finally, we obtain the $b^{th}$ channel's influence on the $a^{th}$ channel (i.e., the channel attention map $\mathcal{M}_c^{a,b}$) by performing the *Softmax* operation as follows:

$$\mathcal{M}_c^{a,b} = Softmax\left(\frac{f(F_a, F_b)}{\sqrt{C}}\right) \quad (5)$$

where $f$ is used to calculate the interdependence between channels. Additionally, we employ global average pooling to capture the channel-wise global discriminatory features with high-level semantic information. The pooling output, being passed through a $1 \times 1$ convolution, is multiplied with $M_c$ and reshaped to a dimension of $1 \times 1 \times C$. Therefore, the final channel-attention output $O^c$ can be obtained as:

$$O_a^c = \zeta \sum_{k=1}^{C}[M^{a,b}{}_c \cdot G(F_i)] + wF_a; \forall i \in \{1,2,...,C\} \quad (6)$$

where G represents global average pooling, $\zeta$ is a tunable parameter, and $w$ is the weight of the $1 \times 1$ convolution.

Spatial Attention: The output of the U-Net encoder passes through the spatial attention branch of the bilinear fusion block (please refer to Section IA in the supplementary material for details). Following the concatenation of obtained channel attention and spatial attention map, a fusion of features from the transformer and U-Net encoder is obtained.

Finally, if $W_1$ and $W_2$ are the weights for the two respective branches of the bilinear fusion block, we obtain a Hadamard product of the input features, which can be expressed as in Equation 7.

$$B_i = Conv(Z_i W_{1i} \odot U_i W_{2i}) \quad (7)$$

where $\odot$ represents the Hadamard operation, $Z_i$ and $W_i$ represent the input features from the two branches. Finally, $B_i$ is concatenated with the outputs from the spatial and channel attention branches, forming the final output of the bilinear fusion block. The spatial attention block is shown in Figure 2.

### D. Multi-Scale Context Extraction

In the proposed architecture, RFB is the core component of the context extractor module, placed in the skip connections of U-Net. As shown in Figure 2, the RFB blocks receive context information from multiple downsampling branches with varying scales making the network adaptable to multiscale context information. Dilated convolution in pairs of ($n \times 1$ and $1 \times n$) is employed for each branch unlike in [25]. For further details please refer to Section IB of the supplementary material. Thus, incorporating the stated arrangement of differently sized kernels in the branches acts superior to the fixed-sized receptive fields, enhancing the overall feature extraction for further reconstruction. The decoder path used for the up-sampling operation is explained in Section IC of supplementary material.

### E. Loss function

This work employs a linear combination of dice coefficient, Intersection over Union (IoU), and Binary CrossEntropy (BCE) loss for training the network. The proposed segmentation pipeline utilizes the ground truth for supervision of the overall segmentation procedure. The BCE loss examines each pixel individually asserting equal learning to each pixel without blowing up the gradient while the Dice coefficient measures the extent of overlap between the ground truth and a segmentation map. The said BCE loss, Dice loss, and IoU loss can be expressed as in Equation 8, Equation 9, and Equation 10 respectively:

$$\mathcal{L}_{BCE}(GT,S) = \sum_j -(GT_i \cdot \log S_i + (1-GT_i) \cdot \log(1-S_i)) \quad (8)$$

$$\mathcal{L}_{Dice}(GT,S) = 1 - \sum_{k=1}^{c} \frac{2\omega_k \sum_{j=1}^{n} S(k,j)GT(k,j)}{\sum_{j=1}^{n} S(k,j)^2 + \sum_{j=1}^{N} GT(k,j)^2} \quad (9)$$

$$\mathcal{L}_{IoU}(GT,S) = 1 - IoU(GT,S) - \frac{|X - GT \cup S|}{|X|} \quad (10)$$

where *GT* is the ground truth, *S* refers to the prediction map obtained after segmentation, $GT_i$ and $S_i$ are a single image from the ground truth labels and predicted labels respectively, where $i \in \{1,2,3,...,N\}$ and $k \in \{1,2,3,...,c\}$, where *N* represents the number of image data available and *n* represents the pixel number. $\omega_k$ is the weight of the $k^{th}$ class. Variable *X* denotes the smallest bounding box covering the segmentation map and the ground truth. Therefore, the overall loss function can be mathematically represented as shown in Equation 11.

$$L(GT,S) = L_{MAE}(GT,S) + \lambda_1 L_{Dice}(GT,S) \quad (11)$$
$$+ \lambda_2 L_{IoU}(GT,S) + \lambda_3 L_{BCE}(GT,S)$$

$\lambda_1$, $\lambda_2$, $\lambda_3$, and $\lambda_4$ hyperparameters are experimentally set to 0.15, 0.4, 0.3, and 0.15 respectively. $L_{MAE}$ denotes mean absolute error loss. For detailed analysis, please refer to Section IV of the supplementary material.

The proposed method is quite efficient for accurate segmentation of optic disc and optic cup from retinal images, as it utilizes the advantages of both the CNN and transformer network. The accurate segmentation performance can therefore be utilized for the measurement of CDR to indicate the onset of Glaucoma. In the following section, we discuss the experimental results in detail. We also compare the performance and inference time with several other state-of-the-art methods in the literature.

## III. EXPERIMENTAL RESULTS

The proposed UT-Net has been experimented on three publicly available datasets, namely DRISHTI-GS [2], RIM-ONE R3 [26], and REFUGE [27]. Five metrics: Dice Similarity Coefficient (DSC), IoU, Precision, Sensitivity, and Accuracy are used to evaluate the performance. The implementation details are provided in Section II of the supplementary material.

TABLE I: Quantitative results of the proposed UT-Net on the three publicly available datasets for OD and OC segmentation. DSC: Dice Similarity Coefficient, IoU: Intersection over Union

| Dataset | Type | DSC | IoU | Precision | Sensitivity | Accuracy |
|---|---|---|---|---|---|---|
| DRISHTI-GS [2] | Optic Disc | 97.92 | 94.22 | 96.52 | 97.19 | 99.84 |
|  | Optic Cup | 94.09 | 89.67 | 94.95 | 96.48 | 99.84 |
| RIM-ONE R3 [26] | Optic Disc | 96.14 | 91.87 | 94.71 | 97.51 | 99.87 |
|  | Optic Cup | 92.49 | 84.81 | 93.54 | 97.67 | 99.83 |
| REFUGE [27] | Optic Disc | 95.93 | 90.50 | 94.48 | 95.75 | 99.84 |
|  | Optic Cup | 92.67 | 89.96 | 91.44 | 91.44 | 99.83 |

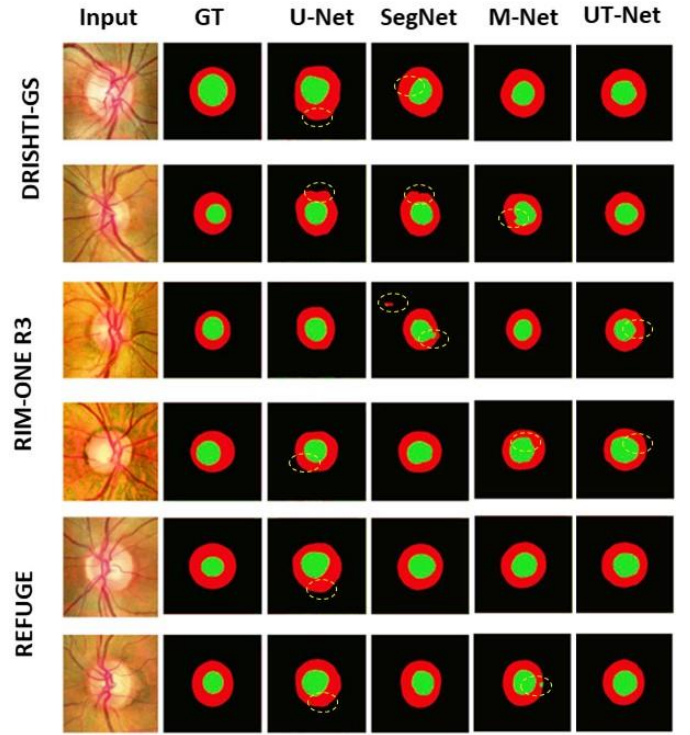

Fig. 4: Visual comparison of the results obtained by the proposed UT-Net on three datasets with the existing state-of-the-art methods. The red and green regions represent the optic disc and optic cup respectively. The highlighted regions represent the wrongly segmented parts.

For all the metrics, the average on all the validation set images is considered for evaluation. Some precise and promising OC and OD segmentation performances are obtained on these three datasets, as shown in Table I. The viability of the proposed UT-Net can be justified by the high values of mean DSC and IoU measures. One interesting observation from Table I is that the segmentation performance on OC segmentation is always lower than that on OD segmentation. This is due to the obscure boundary of a very small OC region inside the OD, as shown in Figure 1. Still, UT-Net produces commendable results on OC segmentation, adding to the robustness of the proposed method.

### A. Results on the DRISHTI-GS dataset

When evaluated on the DRISHTI-GS dataset for OD segmentation, our model produces scores of 97.92%, 94.22%, 97.19%, and 99.84% in terms of DSC, IoU, sensitivity, and accuracy respectively. Similarly, when evaluated on the OC segmentation task, it produces scores of 94.09%, 89.67%, 96.42%, and 99.84% in respective metrics. Table I depicts the complete result on OD and OC segmentation using the five evaluation metrics, as discussed before. When compared to the state-of-the-art methods in the literature, UT-Net performed quite well, producing the best results in OD segmentation in terms of DSC, IoU, and accuracy. Son et al. [31] produced the best sensitivity result in OD segmentation but performed quite poorly in terms of other metrics. Table II shows a detailed quantitative comparison of our results with others. When compared to the OC segmentation task, our method outperforms state-of-the-art methods like M-Net [29], CENet [12], DRNet [13], etc. by a significant margin. The first two rows of Figure 4 show the visual comparison of our segmentation performance with U-Net [22], SegNet [33], and M-Net [29] where the OD and OC are marked in red and green respectively. It is clear that despite being highly efficient in several segmentation tasks, U-Net and SegNet produce substandard results on this particular task. Although M-Net results are close to ours, the proposed UT-Net generates the best performance among all.

TABLE II: Quantitative comparison of the results obtained by the UT-Net and the past methods on the DRISHTI-GS dataset for OD and OC segmentation.

| Method | Optic Disc | | | | Optic Cup | | | |
|---|---|---|---|---|---|---|---|---|
| | DSC | IoU | Sensitivity | Accuracy | DSC | IoU | Sensitivity | Accuracy |
| CDED-Net [11] | 95.97 | 91.83 | 97.54 | 99.73 | 92.4 | 86.32 | 95.61 | 99.81 |
| CE-Net [12] | 96.42 | 93.23 | 97.59 | 99.81 | 88.18 | 80.06 | 88.19 | 99.09 |
| pOSAL [28] | 96.5 | - | - | - | 85.8 | - | - | - |
| M-Net [29] | 96.78 | 93.86 | 97.11 | 99.91 | 86.18 | 77.3 | 88.22 | 98.62 |
| Lei et al. [30] | 96.58 | 93.33 | 97.18 | - | 89.21 | 87.98 | 96.15 | - |
| DRNet [13] | - | 93.3 | 96 | 99.85 | - | - | - | - |
| Son et al. [31] | 95.27 | 91.85 | 97.47 | 99.71 | 86.43 | 77.48 | 85.39 | 99.07 |
| Sevastopolsky et al. [10] | 90.43 | 83.50 | 91.56 | 99.69 | 85.21 | 75.15 | 84.76 | 98.81 |
| Xu et al. [32] | 97.84 | 92.93 | - | - | 89.20 | 82.30 | - | - |
| UT-Net (Proposed) | 97.92 | 94.22 | 97.19 | 99.84 | 94.09 | 89.67 | 96.48 | 99.84 |

TABLE III: Quantitative comparison of the results obtained by the UT-Net and the past methods on the RIM-ONE R3 dataset for OD and OC segmentation.

| Method | Optic Disc | | | | Optic Cup | | | |
|---|---|---|---|---|---|---|---|---|
| | DSC | IoU | Sensitivity | Accuracy | DSC | IoU | Sensitivity | Accuracy |
| CDED-Net [11] | 95.82 | 91.01 | 97.34 | 99.73 | 86.22 | 75.32 | 95.17 | 99.81 |
| CE-Net [12] | 95.27 | 91.15 | 95.02 | 99.86 | 84.35 | 74.24 | 83.52 | 99.70 |
| pOSAL [28] | 86.5 | - | - | - | 78.7 | - | - | - |
| M-Net [29] | 95.26 | 91.14 | 94.81 | 99.86 | 83.48 | 73.00 | 81.46 | 99.67 |
| Sevastopolsky et al. [10] | 95.00 | 82.00 | 95.09 | 99.73 | 82.00 | 69.00 | 75.45 | 99.76 |
| Lei et al. [30] | 90.76 | 88.41 | 95.21 | - | 82.18 | 83.73 | 97.54 | - |
| DRNet [13] | 94.12 | 90.1 | 95.9 | 96.2 | - | - | - | - |
| Son et al. [31] | 95.32 | 91.22 | 94.57 | 99.87 | 92.50 | 71.67 | 81.42 | 99.65 |
| Xu et al. [32] | 95.61 | 91.72 | 95.21 | 99.87 | 85.64 | 75.87 | 85.15 | 99.71 |
| UT-Net (Proposed) | 96.14 | 91.87 | 97.51 | 99.87 | 92.49 | 84.81 | 97.67 | 99.83 |

### B. Results on the RIM-ONE R3 dataset

The same metrics, as discussed before, are also implemented to evaluate the performance on the RIM-ONE R3 dataset. As shown in Table I UT-Net produces scores of 96.14%, 91.52%, 96.21%, and 99.87% in DSC, IoU, sensitivity, and accuracy measures respectively on OD segmentation. Coming to the OC part, the measures of DSC and IoU are 92.49% and 84.81% respectively. When compared to the state-of-the-art, as shown in Table III, UT-Net produces marginally better results than all in OD segmentation and excels others significantly in OC segmentation. Recently proposed methods like CDEDNet [11], Xu et al. [32], and Son et al. [31] consistently produced comparable results as ours in OD segmentation. However, they produced relatively worse results in the case of OC segmentation, leaving the scope for further research. The third and fourth rows of Figure 4 show the visual comparison of the results obtained from UT-Net with U-Net, M-Net, SegNet, and corresponding ground truths. In this case, also, U-Net and SegNet generate some uncanny results, for example, in the first image, some of the clustered pixels of the nerve vessel are misclassified as OD.

Additionally, we compare the inference time of our proposed method on the RIM-ONE R3 dataset with several state-of-the-art methods, as shown in Table V. It is to be noted that training a deep learning model for multiple times is not required in medical applications, hence the inference speed is of more importance than the training time, especially in this case. The inference time includes producing the segmentation map, computing CDR, and predicting Glaucoma (refer to subsection III-E for CDR). We compute the inference time for all the images in the test set and calculate the mean inference time for a single image from that. It is evident from Table V that, our proposed method has almost comparable inference time as compared to the SOTA, but it can generate a segmentation map with much higher accuracy. Methods like comparison of our proposed method with several other state-of-the-art methods on the RIM-ONE R3 dataset.

TABLE IV: Quantitative comparison of the results obtained by the UT-Net and the past methods on the REFUGE dataset for OD and OC segmentation.

| Method | Optic Disc | | Optic Cup | |
|---|---|---|---|---|
| | DSC | IoU | DSC | IoU |
| U-Net [22] | 93.08 | 87.50 | 85.44 | 77.60 |
| Mask RCNN [14] | 90.45 | 85.47 | 82.36 | 75.21 |
| SegNet [33] | 92.78 | 88.16 | 87.47 | 79.06 |
| M-Net [29] | 93.59 | 88.26 | 86.48 | 78.97 |
| ET-Net [34] | 95.29 | - | 89.12 | - |
| Efficient U-Net++ [35] | 95.84 | 88.2 | 91.81 | 87.54 |
| pOSAL [28] | 95.50 | - | 85.80 | - |
| Wang et al. [36] | 93.42 | 87.96 | 91.62 | 85.65 |
| CDED-Net [11] - | - | 88.37 | - | 81.11 |
| UT-Net (Proposed) | 95.93 | 90.50 | 92.67 | 89.96 |

TABLE V: Performance against inference time (in milliseconds)

TABLE VI. As observed the U-Net encoder on its own fails to perform well despite its string representation power. This claim is further supported by the OC segmentation where DSC and IoU values of 87.51% and 83.82% are obtained. Transformers on the other hand capture the global dependencies but fail to capture essential semantic information obtained from CNN layers as shown in Table VI.

TABLE VI: Results of the ablation study performed on the DRISHTI-GS dataset to justify the importance of individual architectural components of the proposed UT-Net

| Method | Optic Disc | | | | Optic Cup | | | |
|---|---|---|---|---|---|---|---|---|
| | DSC | IoU | Sensitivity | Accuracy | DSC | IoU | Sensitivity | Accuracy |
| Transformer encoder only (like Chen et al. [7]) | 88.54 | 81.36 | 89.43 | 97.87 | 84.52 | 77.14 | 89.61 | 96.24 |
| U-Net encoder (like Ronneberger et al. [22]) | 90.43 | 84.33 | 92.04 | 98.42 | 87.51 | 83.82 | 90.40 | 97.73 |
| Transformer+U-Net encoder (simple concatenation) | 94.93 | 89.36 | 95.11 | 99.01 | 90.14 | 86.26 | 93.76 | 98.88 |
| Transformer+U-Net encoder (bilinear fusion) | 96.44 | 92.25 | 96.73 | 99.42 | 93.36 | 88.96 | 95.16 | 99.21 |
| UT-Net (Proposed) | 97.92 | 94.22 | 97.19 | 99.84 | 94.09 | 89.67 | 96.48 | 99.84 |

| Model | DSC (OD) | DSC (OC) | Inference Time (millisec.) |
|---|---|---|---|
| U-Net [22] | 86.77 | 78.32 | 90.89 |
| CE-Net [12] | 95.27 | 84.35 | 105.60 |
| M-Net [29] | 94.26 | 83.48 | 110.34 |
| Sevastopolsky et al. [10] | 95.00 | 82.00 | 98.87 |
| pOSAL [28] | 86.50 | 78.70 | 112.12 |
| U-Net++ [18] | 89.45 | 80.44 | 108.34 |
| UT-Net (Proposed) | 96.14 | 92.49 | 111.51 |

[10], [22], although has a faster inference speed, they compromise with the segmentation accuracy. Our model, on the other hand, maintains a trade-off between computation and performance.

### C. Results on the REFUGE dataset

As shown in Table I, the results obtained on the REFUGE dataset are 95.93%, 90.50%, 95.75%, and 99.84% in terms of DSC, IoU, sensitivity, and accuracy respectively in OD segmentation, and 92.67%, 89.96%, 91.44%, and 99.83% in OC segmentation in terms of the same metrics.
For comparison with state-of-the-art methods like U-Net [22], Mask RCNN [14], ET-Net [34], etc., we only use DSC and IoU in Table IV because of the unavailability of other metrics for comparison. It is clear from the table that recent developments like pOSAL [28], and Efficient U-Net++ [35] also produced comparable results with UT-Net, both in OC and OD segmentation, but marginally better results of UT-Net make it the topper than all of them. The last two rows of Figure 4 depict the visual comparison of some of these methods with UT-Net. Small and inconspicuous OC regions are so difficult to segment that state-of-the-art methods like M-Net often misclassify some of the OD regions as OC. Compared to that our proposed UT-Net consistently produces superior results.

### D. Ablation Study

To observe and justify the effectiveness of multiple encoders, bilinear fusion, and context extractor modules, we perform an ablation study on the DRISHTI-GS dataset, as discussed below. The experimental results are depicted in Table VI. The combination of the encoders effectively improves the feature representation in the space.

These extracted features when combined via the bilinear fusion block produce appreciable results of 96.44% and 93.36% DSC for OD and OC respectively distinguishing between discriminative and redundant features. Finally, the importance of the multi-scale extraction of features for reconstruction is shown in the last row of Table VI. We have provided a detailed analysis and contribution of every component used in the ablation study in Section III of the supplementary file.

TABLE VII: Comparison of CDR errors ($\delta_{CDR}$) for different methods on the three datasets. A lower value indicates better performance.

| Method | DRISHTI-GS | RIM-ONE R3 | REFUGE |
|---|---|---|---|
| Hervella et al. [37] | 0.0413 | - | 0.0373 |
| Sun et al. [38] | 0.0499 | 0.063 | - |
| Transformer encoder + decoder | 0.0624 | 0.0791 | 0.0557 |
| U-Net encoder + decoder | 0.0488 | 0.0711 | 0.0441 |
| UT-Net (Proposed) | 0.0399 | 0.0617 | 0.0368 |

### E. Glaucoma Detection

The CDR is an important and widely used parameter by ophthalmologists for the screening of glaucoma. It can be expressed as:

$$CDR = \frac{D_c}{D_d} \quad (12)$$

where $D_c$ and $D_d$ represent vertical cup diameter and vertical disc diameter, respectively. Based on the obtained CDR values, a patient can be diagnosed as normal ($CDR \leq 0.4$), mild glaucomatous ($CDR \in \{0.4, 0.5\}$), and severe glaucomatous ($CDR \in \{0.5, 0.8\}$). We assess the performance of our proposed model in the accurate detection of glaucoma by calculating the CDR error value $\delta_{CDR}$, which can be defined as:

$$\delta_{CDR} = \frac{1}{N} \sum_{n=1}^{N} |CDR_{\mathcal{G}}^n - CDR_{\mathcal{O}}^n| \quad (13)$$

where $CDR_G$ is the CDR score computed on ground truth annotation and $CDR_O$ denotes the obtained CDR value, $N$ is the total number of samples in the test set. A low value of $\delta_{CDR}$ indicates better performance for Glaucoma detection. We obtain $\delta_{CDR}$ values of 0.0399, 0.0617, and 0.0368 for DRISHTI-GS, RIM-ONE R3, and REFUGE datasets, respectively. Table VII refers to the comparison of $\delta_{CDR}$, evaluated on the three datasets used with other methods. It is visible that the proposed method not only improves the segmentation performance as compared to the standalone transformer and U-Net models but also observes a prominent improvement in CDR measurement, i.e., Glaucoma detection. Besides, we also observe a significant improvement in performance (i.e., low $\delta_{CDR}$ value) over the methods in the literature, adding to the clinical significance of our proposed method.

We demonstrate the ROC curve for the segmentation of OD and OC in Figure 5 (a)-(c), where it is visible that OD segmentation performance is better as compared to OC segmentation.

TABLE VIII: Comparison of our proposed method with some existing methods in terms of AUC score obtained on the DRISHTI-GS dataset

| Author | Method | Mean AUC |
|---|---|---|
| Perdomo et al. [39] | Morphometric Feature Estimation | 0.82 |
| Chakravarty et al. [40] | Deep CNN Features | 0.881 |
| Hervella et al. [37] | Multi task learning | 0.947 |
| Orlando et al. [41] | Vessel inpainting | 0.863 |
| CDED Net [11] | SegNet like architecture | 0.963 |
| UT-Net (Proposed) | Transformer+U-Net fusion | 0.984 |

We also compute the Area Under the Curve (AUC) to assess the performance of our proposed method. The obtained mean AUC scores are 0.984, 0.957, and 0.958 on DRISHTI-GS, RIM-ONE R3, and REFUGE datasets, respectively. A high AUC score indicates a high percentage of correctly classified images by the system, justifying its robustness for Glaucoma detection. As shown in Table VIII, our proposed method achieves the best results in terms of AUC scores as compared to several existing methods in the literature. Apart from this, the accuracy of CDR measurement is also computed in terms of Pearson Correlation Coefficient (PCC) values between the measured CDR and the actual CDR of annotation. The CDR measurement performance of the proposed method with respect to the ground truth annotation is shown in Figure 5 (d)(f) for all three datasets. The proposed method achieves PCC scores of 91.2%, 89.1%, and 88.5% on DRISHTI-GS, RIMONE R3, and REFUGE datasets, justifying the effectiveness of our proposed method for accurate glaucoma detection.

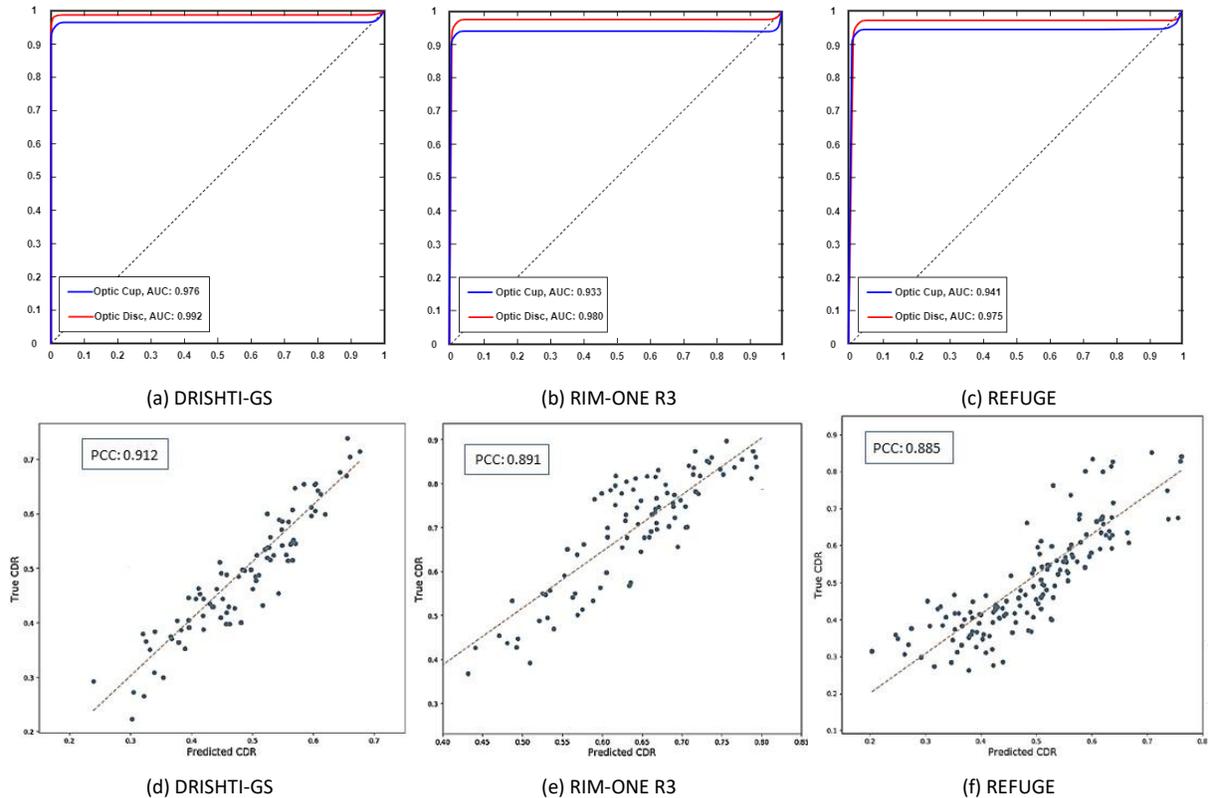

Fig. 5: ROC curve for segmentation of OD and OC on (a) DRISHTI-GS, (b) RIM-ONE R3, and (c) REFUGE datasets. Performance evaluation of CDR measurement in terms of scatter plot for the three datasets along with their PCC values for (d) DRISHTI-GS, (e) RIM-ONE R3, and (f) REFUGE datasets.

## IV. CONCLUSION AND FUTURE WORK

Approximately 3.54% of the world's population is estimated to be glaucomatous. Hence, early detection of glaucoma is essential for proper treatment. To assist clinicians in facilitating the computer-aided diagnosis of the cause, this paper proposes an efficient deep-learning framework, called UT-Net. With this, we address several shortcomings of the existing frameworks. UT-Net combines the ability of a UNet encoder for acquiring high-level semantic information with a transformer encoder that can compensate for the localization problem of the CNNs, extracting global contextual information. Furthermore, the skip connections of U-Net are kept intact to recover the high-level information from the UNet encoder. Besides, context extractors are introduced in the skip connections to amalgamate receptive field information at multiple levels. The proposed method is evaluated on three publicly available datasets, DRISHTI-GS, RIM-ONE R3, and REFUGE for segmentation of optic disc and cup, and evaluated for Glaucoma detection. The model outperforms several state-of-the-art architectures by a significant margin. The ability of the model to distinguish between the low contrast region of OD and OC shows noticeable improvements in the calculation of CDR, thus detecting glaucoma accurately. The proposed architecture at its current state can effectively be used as a diagnostic tool for better treatment. As Glaucoma is a major concern of the 21st century, heuristic approaches cannot always be trusted for large-scale screening. Our proposed method can be useful to this end, to effectively map the risk profile of multiple patients that can help physicians with better treatment planning. Our method is non-invasive and utilizes fundus images, which are comparatively cheaper to obtain as compared to traditional OCT images. The segmentation results can be used to calculate the CDR indicating the onset of glaucoma. The accuracy of the CDR is directly dependent on the output segmentation maps demonstrating the model's utility in the overall CAD pipeline. Additionally, several minor details are often overlooked even by the expert human eyes, which we can capture with precision using the CAD models like ours. Therefore OD and OC segmentation play a huge role in the inspection of glaucomatous cupping.

In the future, we would like to extend our work to several other segmentation tasks in Retinopathy for instance exudates and lesion segmentation in fundus images. We also plan to incorporate adversarial learning similar to [30] to address the unavailability of large-sized datasets in the domain. Fuzzy fusion [42] of multiple encoded features can be an option to improve the representational ability of the segmentation frameworks.